\pdfoutput=1

\documentclass[11pt]{article}

\usepackage[preprint]{acl}

\usepackage{times}
\usepackage{latexsym}

\usepackage[T1]{fontenc}

\usepackage[utf8]{inputenc}

\usepackage{microtype}

\usepackage{inconsolata}

\usepackage{graphicx}

\usepackage{enumitem}
\usepackage{algorithm}
\usepackage{algpseudocode}
\usepackage{xcolor}
\algrenewcommand\algorithmiccomment[1]{\hfill\textcolor{blue}{$\triangleright$ #1}}
\usepackage{amsmath}
\usepackage[normalem]{ulem} 
\newcommand{\redst}[1]{\textcolor{red}{\sout{#1}}}
\usepackage{booktabs} 
\usepackage{tcolorbox}
\usepackage{subfigure}
\usepackage{caption}
\title{\textsc{TreeCut}: A Synthetic Unanswerable Math Word Problem Dataset for LLM Hallucination Evaluation}

\author{Jialin Ouyang \\
  Columbia University \\
  \texttt{jo2559@columbia.edu}}

\begin{document}
\maketitle
\begin{abstract}

Large language models (LLMs) now achieve near-human performance on standard math word problem benchmarks (e.g., GSM8K), yet their true reasoning ability remains disputed. A key concern is that models often produce confident, yet unfounded, answers to unanswerable problems. We introduce \textsc{TreeCut}, a synthetic dataset that systematically generates \emph{infinite} unanswerable math word problems and their answerable counterparts, by representing each question as a tree and removing chosen necessary conditions. Experiments show \textsc{TreeCut} effectively induce hallucinations in large language models, including GPT-4o and o3-mini, with rates of 64\% and 44\% in their respective worst-case scenarios under zero-shot setting.
Further analysis highlights that deeper or more complex trees, composite item names, and removing necessary condition near the middle of a path all increase the likelihood of hallucinations, underscoring the persistent challenges LLMs face in identifying unanswerable math problems. The dataset generation code and sample data are available at \url{https://github.com/j-bagel/treecut-math}.

\end{abstract}

\begin{figure*}[t!]
    \centering
    \begin{minipage}[t]{0.25\textwidth}  
        \vspace{0pt}  
        \includegraphics[width=\textwidth]{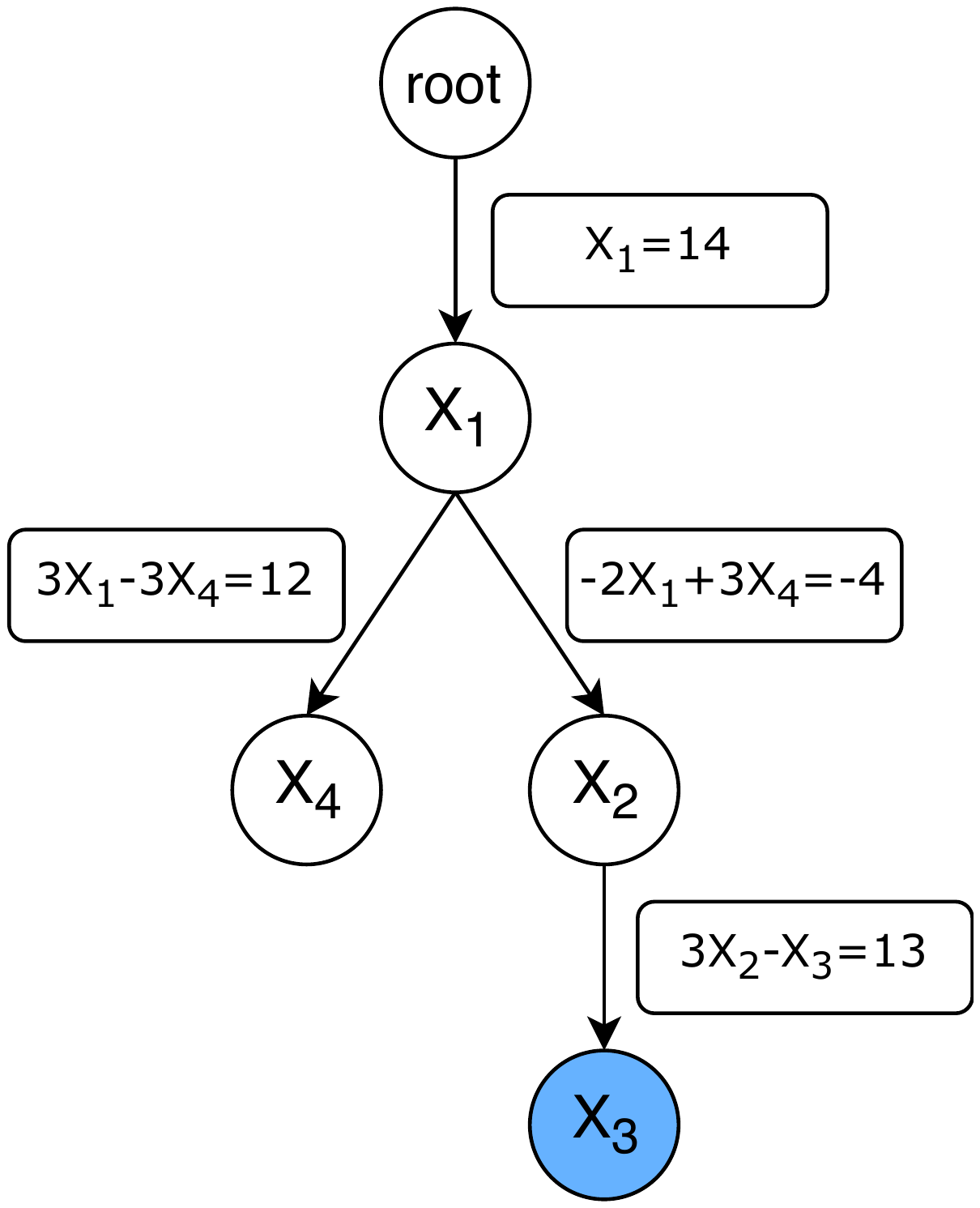}
    \end{minipage}%
    \begin{minipage}[t]{0.25\textwidth}  
        \vspace{0pt}  
        \includegraphics[width=\textwidth]{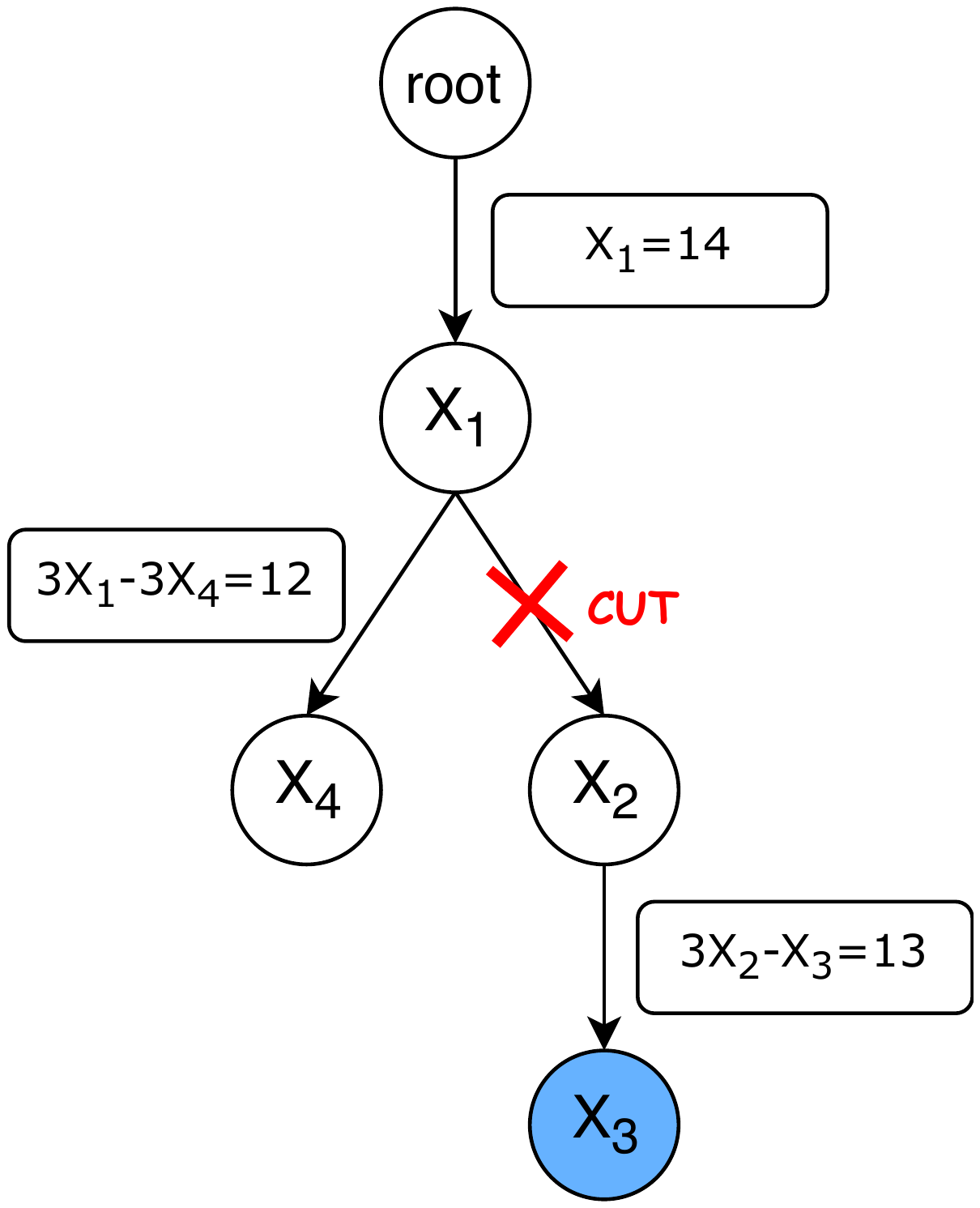}
    \end{minipage}%
    \hspace{0.3em}
    \begin{minipage}[t]{0.46\textwidth}  
        \vspace{1pt}  
        \fcolorbox{black}{gray!20}{%
            \parbox{\textwidth}{%
                \scriptsize
                \textbf{Question:} \\
                A burger costs 14 dollars. \redst{3 scrambled eggs cost 4 dollars less than 2 burgers.} 3 pies cost 12 dollars less than 3 burgers. A BLT sandwich costs 13 dollars less than 3 scrambled eggs. Question: how much does a BLT sandwich cost? \\ \\
                \textbf{Solution to the answerable problem:} \\
                It is given as a fact that a burger costs 14 dollars. Combine with the fact that 3 scrambled eggs cost 4 dollars less than 2 burgers, we get a scrambled egg costs 8 dollars. Combine with the fact that a BLT sandwich costs 13 dollars less than 3 scrambled eggs, we get a BLT sandwich costs 11 dollars. \\ \\
                \textbf{Solution to the unanswerable problem:} \\
                All we know about the prices of BLT sandwich and scrambled egg is: a BLT sandwich costs 13 dollars less than 3 scrambled eggs.
                There are 2 variables but only 1 linear formula, so we cannot calculate the price of a BLT sandwich.
            }
        }
    \end{minipage}
    \caption{The left and middle panels depict the tree structures corresponding to the answerable and unanswerable questions, respectively. In the right panel, the strike-through sentence represents the formula removed by the \textit{cut}. The variable mappings to items are as follows: $x_1$ represents a burger, $x_2$ represents a scrambled egg, $x_3$ represents a BLT sandwich, and $x_4$ represents a pie.}
    \label{f:tree1}
\end{figure*}

\section{Introduction}
\label{sec:introduction}

Mathematical reasoning is a crucial part of human intelligence. Recent years have witnessed remarkable advancements in the mathematical reasoning capabilities of large language models (LLMs).  By leveraging techniques such as chain-of-thought prompting~\citep{wei2022chain}, state-of-the-art LLMs (e.g., \citet{achiam2023gpt, team2024gemini, dubey2024llama}) achieved human-level performance on benchmarks like GSM8K~\citep{cobbe2021training}. However, it  remains controversial whether this performance implies reasoning capability beyond pattern matching.

A substantial body of research highlights the capability of Large Language Models in mathematical reasoning.
\citet{achiam2023gpt, team2024gemini, dubey2024llama, yang2024qwen2}, among others, achieved over 90\% accuracy on GSM8K~\citep{cobbe2021training}, a dataset consists of 8K grade school math word problems. \citet{yang2024qwen2, zhou2023solving}, among others, achieved over 80\% accuracy on the more difficult MATH dataset~\citep{hendrycks2021measuring}, which consists of 12.5K high school math competition problems.

Meanwhile, there is a line of research questioning the reasoning ability of LLMs by showing their vulnerability under superficial changes of the input that do not alter the underlying logic.
Works like \citet{shi2023large, jiang2024peek} find that LLMs are easily distracted by irrelevant context or token level perturbation that does not change the underlying logic of the reasoning task. \citet{mirzadeh2024gsm} further demonstrate that the performance of LLMs declines when numerical values are altered in the questions from the GSM8K dataset.

There is yet another line of research that challenges the ability of LLMs to refrain from answering unanswerable problems. \citet{ma2024large, li2024gsm, sun2024benchmarking, zhou2024your, saadat2024not} introduce minor modifications to existing math word problems to create unanswerable variants, and find that LLMs often generate hallucinatory answers for these unanswerable questions, even when they perform well on the original answerable datasets. However, these efforts rely on pre-existing math word problem sources, making them susceptible to training data contamination, limited in scope, and lacking rich structures for extended research.

To address these shortcomings, we propose \textsc{TreeCut}, a synthetic dataset capable of systematically generating an infinite number of unanswerable math word problems and their answerable counterparts. \textsc{TreeCut} considers problem represented by a tree, with nodes representing variables and edges representing formulas. Unanswerable problems are generated by removing an edge along the path from the root to the questioned variable. Our unanswerable dataset proves to be challenging even for GPT-4o and o3-mini. In addition, \texttt{TreeCut} allows precise control over the structural components of each problem, enabling detailed investigations into when and why LLMs produce hallucinations. Our analysis highlights that deeper or more complex trees, composite item names, and removing necessary condition near the middle of a path all increase the likelihood of hallucinations.

\section{Related Work}
\paragraph{Math Word Problem Benchmark}
Numerous math word problem datasets of different difficulty have been proposed in previous research, most notable examples including GSM8K \citep{cobbe2021training} and MATH \citep{hendrycks2021measuring}.

Many benchmarks have been developed to measure the robustness of mathematical reasoning. \citet{patel2021nlp, kumar2021adversarial, xu2022towards, li2024gsm, zhou2024mathattack, yu2023metamath, shi2023large} perturb or rewrite math word problems to measure the robustness of mathematical reasoning. 

\citet{liu2021roda} utilize tree structures to represent and manipulate mathematical expressions during the reverse operation based data augmentation process for MWP solving.
\citet{opedal2024mathgap} introduced MathGAP, a framework for evaluating LLMs using synthetic math word problems
with controllable proof tree characteristics. 
In contrast to their approach, the tree structure in our problem-generation procedure is fundamentally different. In our work, each node represents a variable, and the questioned variable appears as a leaf. In their work, however, each node represents a logical statement, with the answer represented by the root. More importantly, we focus on unanswerable math word problems, an aspect that their study did not address.

\paragraph{Unanswerable Math Problems}

\citet{yin2023large} introduced SelfAware, consisting
of unanswerable questions from five diverse categories. It includes less than 300 unanswerable mathematical problems. \citet{li2024gsm, zhou2024your} generate unanswerable questions by prompting GPT4 to eliminate a necessary condition from the original problem, an then the modified questions are further checked or refined by human annotators. \citet{ma2024large} prompt GPT4 to modify problems from GSM8K. 
\citet{sun2024benchmarking} task human annotators to modify original questions in existing MWP datasets to make them unanswerable, creating a dataset composed of 2,600 answerable questions and 2,600 unanswerable questions.

\begin{table*}[t!]
    \centering
    \small
    \renewcommand{\arraystretch}{1} 
    \begin{tabular}{c|ccccccc}
        \toprule
        \texttt{ansDepth} & \textbf{Llama-8B} & \textbf{Llama-70B} & \textbf{Qwen-7B} & \textbf{Qwen-72B} & \textbf{GPT-4o} & \textbf{o3-mini} \\
        \midrule
2  &  80.2\%  &  24.6\%  &  84.6\%  &  59.8\%  &  12.0\%  &  44.0\% \\
4  &  86.2\%  &  40.2\%  &  90.4\%  &  82.8\%  &  18.0\%  &  25.2\% \\
6  &  86.0\%  &  63.4\%  &  95.6\%  &  88.4\%  &  47.4\%  &  19.2\% \\
8  &  84.2\%  &  65.0\%  &  93.4\%  &  85.2\%  &  64.0\%  &  25.6\% \\
        \bottomrule
    \end{tabular}
    \captionsetup{skip=3pt}
    \caption{\small Percentage of hallucination of various LLMs at different \texttt{ansDepth} values for unanswerable problems, zero-shot prompting}
    \label{tab:unknown}
\end{table*}

\begin{table*}[t]
    \centering
    \small
    \renewcommand{\arraystretch}{1} 
    \begin{tabular}{c|ccccccc}
        \toprule
        \texttt{ansDepth} & \textbf{Llama-8B} & \textbf{Llama-70B} & \textbf{Qwen-7B} & \textbf{Qwen-72B} & \textbf{GPT-4o} & \textbf{o3-mini} \\
        \midrule
2  &  72.8\%  &  33.6\%  &  80.4\%  &  55.4\%  &  18.4\%  &  3.2\% \\
4  &  79.0\%  &  57.6\%  &  94.6\%  &  84.8\%  &  28.8\%  &  2.4\% \\
6  &  79.6\%  &  72.4\%  &  92.6\%  &  84.8\%  &  41.8\%  &  3.6\% \\
8  &  78.6\%  &  76.8\%  &  94.4\%  &  83.0\%  &  51.0\%  &  3.0\% \\
        \bottomrule
    \end{tabular}
    \captionsetup{skip=3pt}
    \caption{\small Percentage of hallucination of various LLMs at different \texttt{ansDepth} values for unanswerable problems, few-shot prompting}
    \label{tab:fewshot}
\end{table*}

\begin{table*}[h]
    \centering
    \small
    \renewcommand{\arraystretch}{1} 
    \label{tab:known}
    \begin{tabular}{c|ccccccc}
        \toprule
        \texttt{ansDepth} & \textbf{Llama-8B} & \textbf{Llama-70B} & \textbf{Qwen-7B} & \textbf{Qwen-72B} & \textbf{GPT-4o} & \textbf{o3-mini} \\
        \midrule
2 & 68\% (14\%) & 95\% (1\%) & 87\% (2\%) & 95\% (1\%) & 99\% (1\%) & 100\% (0\%) \\
4 & 28\% (12\%) & 82\% (6\%) & 31\% (6\%) & 86\% (6\%) & 94\% (0\%) & 100\% (0\%) \\
6 & 17\% (16\%) & 83\% (3\%) & 12\% (9\%) & 80\% (7\%) & 85\% (3\%) & 100\% (0\%) \\
8 & 5\% (12\%) & 76\% (7\%) & 7\% (10\%) & 68\% (8\%) & 84\% (2\%) & 100\% (0\%) \\
        \bottomrule
    \end{tabular}
    \captionsetup{skip=3pt}
    \caption{\small Accuracy of various LLMs at different \texttt{ansDepth} levels for answerable problems. The percentage in parentheses represents the proportion of answerable questions incorrectly identified as unanswerable.}
\end{table*}

\section{\textsc{TreeCut}: a Synthetic (Un)answerable Math Word Problem Dataset}

For the purpose of our investigation, we aim to have full control over the various aspects that determine the underlying \textit{structure} of a math word problem: the name of the entities, the numeric values, and the complexity of the problem. Furthermore, we seek to reliably generate unanswerable problems by precisely removing specific necessary conditions of our choosing.

To this end, we start with a special kind of \textit{answerable} math word problem that can be represented as a tree, as illustrated in Figure \ref{f:tree1}. Within such a tree, each non-root node represents a variable, while the root is a uniquely reserved node. An edge from root gives value to a variable, while an edge between two variables represents a linear formula of the two neighboring nodes. Given such a tree, any variable can be calculated following the unique path from the root to the node that represents the variable. Such a solving procedure does \textit{not} require solving a linear equation system, as the solution only consists of carrying out basic arithmetic operations along the path. To guarantee that the arithmetic operations are well within the capacity of current frontier LLMs,we further restrict the unit price of each food item to be an integer between 5 and 15, and the coefficients of each linear equation taking non-zero integer values between -3 and 3. Finally, variables are randomly mapped to items, and then the formulas are translated to natural language using templates. The complete generation procedure, along with the templates used, is provided in Appendix \ref{sec:algo}.

From an answerable math word problem described above, we generate an unanswerable problem by removing an edge along the path from the root to the questioned variable. In Figure \ref{f:tree1}, $x_3$ is the questioned variable. Along the path to the root, we remove the edge between $x_1$ and $x_2$ (denoted by a \textit{cut}), rendering $x_2$ and $x_3$ undetermined, thus making the question unanswerable, as all we know about $x_2$ and $x_3$ is one single linear equation. A key benefit of such a generation procedure is that the distance from the questioned variable to the \textit{cut} is also fully controlled, as we will see that this factor plays an important role in triggering LLM hallucination.

In summary, we can control the \textit{structure} of problems via the following parameters:
\begin{enumerate}[noitemsep, topsep=0pt, label=-]
\item \texttt{numVars}: total number of variables,
\item \texttt{ansDepth}: distance from the root to the questioned variable,
\item \texttt{compositeName}: boolean, whether the items in the question have composite names (e.g. ``a burger at Bistro Nice'' versus ``a burger''),
\item \texttt{cutDepth}: distance from the questioned variable to the \textit{cut}, if an unanswerable problem is to be generated.
\end{enumerate}
Appendix \ref{sec:algo} contains the detailed problem generation algorithm.

\section{Experiments}
We evaluate several state-of-the-art LLMs using \textsc{TreeCut}. Additionally, we analyze the hallucination rate of GPT-4o on unanswerable problems generated under different parameter configurations of \textsc{TreeCut}.

\subsection{Experimental Setup}
For each set of generation parameters, we randomly generate 500 problems. Unless stated otherwise, we employ a zero-shot prompting template that explicitly directs the model to indicate when a question is unanswerable due to insufficient conditions. A chain-of-thought system message is incorporated for all models except o3-mini\footnote{Following \href{https://platform.openai.com/docs/guides/reasoning}{OpenAI's guidelines} of reasoning models.}. 

\subsection{Evaluating LLMs}
In the first set of experiments, we generate unanswerable math word problems of varying difficulty to evaluate the following LLMs: Llama 3.1 Instruct with 8B and 70B parameters\citep{dubey2024llama}, Qwen2.5 Instruct with 7B and 72B parameters\citep{yang2024qwen2}, GPT-4o\citep{achiam2023gpt}, and o3-mini\citep{openaiO3Mini}. 

Table \ref{tab:unknown} summarizes the results. None of the LLMs gives satisfactory results. Llama 3.1 8B, Qwen2.5 7B and 72B barely have any success identifying unanswerable problems. Llama 3.1 70B and GPT-4o struggle with more complex problems ($\texttt{ansDepth}=6,8$). o3-mini has the lowest hallucination for $\texttt{ansDepth}=6,8$. However, for the easiest case where $\texttt{ansDepth}=2$ (in this setting, only 4 variables are mentioned in each problem), o3-mini displays a bias of making hallucinatory assumptions (see Appendix \ref{sec:o3-mini} for examples).

To further investigate whether the LLMs face intrinsic challenges in recognizing unanswerable math word problems, we conduct another set of experiments using few-shot prompting. For each unanswerable problem, we construct a few-shot prompt by randomly selecting 3 answerable and 3 unanswerable problems, each accompanied by a full solution path and the correct final answer. We use sample size n=500. Results are summarized in Table \ref{tab:fewshot}. O3-mini greatly benefits from few-shot prompting, which is not surprising given our analysis in Appendix \ref{sec:o3-mini}. For shorter problems, o3-mini tends to recognize the lack of conditions during reasoning, but choose to make unreasonable assumptions to arrive at a final answer. Few-shot examples guide it to refrain from doing that. The hallucination rates of the other models remained largely unchanged. This suggests that the five models besides o3-mini face intrinsic challenges in recognizing unanswerable math word problems.

To investigate whether the unsatisfactory accuracy of identifying unanswerable problems comes from the incapability of the necessary mathematical \textit{operations}, we evaluate the LLMs on the \textit{answerable} counterparts of the unanswerable questions using the same zero-shot prompting template. For this experiment, a sample size of $n=100$ is used. We observe that almost every model displays a significant gap between its ability of solving answerable problems and identifying unanswerable problems. For instance, GPT-4o correctly solves 84\% of answerable problems for $\texttt{ansDepth} = 8$, but only correctly recognizes 36\% of unanswerable problems. 

\subsection{Unanswerable Problem Structure and Hallucination}

For a more fine-grained investigation of LLM's hallucination behavior under different \textit{structures} of unanswerable problems, we analyze GPT-4o's hallucination rate on unanswerable problems generated under different parameter choices of \texttt{numVars}, \texttt{ansDepth}, \texttt{compositeName} and \texttt{cutDepth}.

\begin{figure}[h]
    \centering
        \includegraphics[width=0.8\columnwidth]{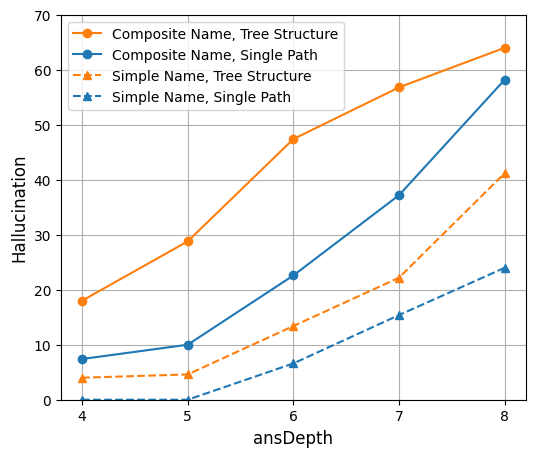}
    \caption{\small Hallucination percentage under different configurations of unanswerable problems, plotted against varying \texttt{ansDepth}.}
  \label{fig:4in1}
\end{figure}

\paragraph{Tree Structure and Item Names} To investigate the effect of (i) a deeper tree structure, (ii) a more complex tree structure, and (iii) composite item names, we consider the following parameter configurations: 
\begin{enumerate}[noitemsep, topsep=0pt, label=$\bullet$]
\item \texttt{ansDepth} $\in \{4,5,6,7,8\}$, which controls the depth of the questioned variable,
\item $\texttt{cutDepth} = \lfloor \texttt{ansDepth} / 2 \rfloor$
\item $\texttt{numVars} = \texttt{ansDepth} + 2$ (generates a more complex tree structure, with conditions unrelated to the questioned variable) or $\texttt{numVars} = \texttt{ansDepth}$ (the tree structure degenerates into a single path),
\item \texttt{compositeName}: $true$ or $false$.
\end{enumerate}
There are $5\times2\times2=20$ configurations in total. We randomly generate 500 unanswerable problems for each configuration, and summarize GPT-4o's hallucination rate in Figure \ref{fig:4in1}. In the figure,
\begin{enumerate}[noitemsep, topsep=0pt, label=-]
\item[$\star$] Orange line represents complex tree structure, 
\item[$\star$] blue line represents simple tree structure,
\item Solid line stands for composite item names, 
\item Dashed line stands for simple item names.
\end{enumerate}

Examining each line individually, we observe that the hallucination rate increases as the depth of the questioned variable grows.
Comparing solid and dashed lines of the same color, a more complex tree structure consistently results in a higher likelihood of hallucination across different \texttt{ansDepth} values.
Comparing orange and blue lines of the same linestyle, composite item names consistently lead to a higher likelihood of hallucination compared to simple item names.

\begin{figure}[h]
    \centering
    \subfigure{
        \includegraphics[width=0.45\columnwidth]{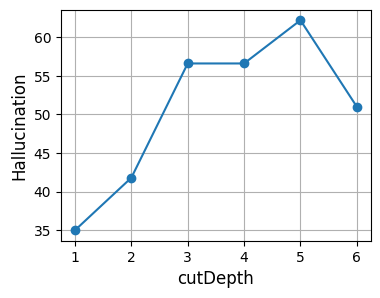}
    }
    \subfigure{
        \includegraphics[width=0.45\columnwidth]{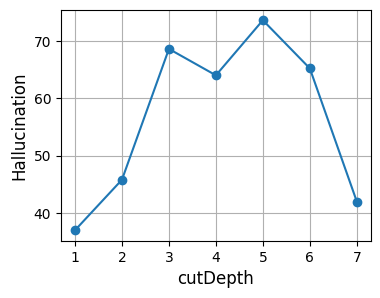}
    }
    \caption{\small Hallucination percentage versus \texttt{cutDepth}. Left panel has $\texttt{ansDepth} = 7$. Right panel has $\texttt{ansDepth} = 8$.}
  \label{fig:cut_exp1}
\end{figure}

\paragraph{Location of the \textit{Cut}} 
For each unanswerable problem, the \textit{cut} always happens along the path from the root to the questioned variable. Does the location of the \textit{cut} change hallucination ratio? We vary \texttt{cutDepth} from 1 to 7 while keeping $\texttt{ansDepth} = 8$ and other parameters fixed. In the right panel of Figure \ref{fig:cut_exp1}, we see that $\texttt{cutDepth}=3,4,5,6$ all trigger over 60\% hallucination for GPT-4o (with $\texttt{cutDepth}=5$ triggering over 70\%), but a $\texttt{cutDepth}=1,2,7$ only triggers less than 50\% of hallucination, which means that GPT-4o is more confused when the \textit{cut} happens around the middle point along the path, comparing to that happens near the root or the questioned variable.

\subsection{Conclusion of Experiments}
Our findings indicate that the unanswerable math word problems generated by \textsc{TreeCut} effectively induce hallucinations in large language models, including GPT-4o and o3-mini, with rates of 61\% and 42\% in their respective worst-case scenarios. Focusing on GPT-4o, we further observe that hallucinations are more likely to occur when the problem exhibits (i) a deeper tree structure, (ii) a more complex tree structure, (iii) composite item names, or (iv) a \textit{cut} positioned around the middle of the path. These results underscore the challenges LLMs face in handling unanswerable math problems.

\section*{Limitations}
Our synthetic dataset is specifically designed for math word problems, representing only a small subset of the broader field of mathematics. Additionally, our evaluations are based on zero-shot and few-shot chain-of-thought prompting. We do not explore alternative prompting techniques commonly used in LLM-based mathematical reasoning studies, which may impact performance comparisons. 

\section*{Acknowledgements}
We thank the anonymous reviewers for their valuable feedback, which helped improve the quality of this work.

\bibliography{main}

\onecolumn
\appendix

\section{Generation of the Math Word Problems}
\label{sec:algo}

{\small
\begin{algorithm}[ht!]
\caption{Generating Math Word Problem using Random Tree}
\label{alg:gen_question}
\begin{algorithmic}[1]
\Require \texttt{numVars} $\geq$ \texttt{ansDepth} $\ge 2$
\Require \texttt{unanswerable} $\in \{true, false\}$, \texttt{order} $\in \{\text{"forward"}, \text{"backward"}, \text{"random"}\}$
\Require \texttt{cutDepth: int}
\If{\texttt{unanswerable} $= true$}
    \Require \texttt{cutDepth: int}, satisfying $1\le$ \texttt{cutDepth} $<$ \texttt{ansDepth}
\EndIf

\Statex \Comment{(i) Sample a dictionary of variable values}
\State \texttt{varDict} $\gets \{\}$
\For{$i \gets 1$ to \texttt{numVars}}
    \State Sample an integer $v \in [5, 15]$
    \State \texttt{varDict}[$x_i$] $\gets v$
\EndFor

\Comment{(ii) Build the random tree}
\State Assign \texttt{root} as the parent of $x_1$
\For{$i \gets 2$ to \texttt{ansDepth}}
    \State Assign $x_{i-1}$ as the parent of $x_i$
\EndFor \Comment{Finish building the path from the root to the questioned variable}
\Statex \Comment{Assign the remaining nodes}
\For{$i \gets \texttt{ansDepth} + 1$ to \texttt{numVars}}
    \State Randomly select a node $x_p$ in the tree
    \State Assign $x_p$ as the parent of $x_i$
\EndFor

\Comment{(iii) Get the list of all edges via a breadth-first traversal}
\State \texttt{edgeList} $\gets$ the list of edges collected by a breadth-first traversal (see Algorithm \ref{alg:tree})

\Comment{(iv) For unanswerable problems, create the \textit{cut}}
\If{\texttt{unanswerable} $= true$}
    \State Remove $(x_{\texttt{ansDepth-cutDepth-1}}, x_{\texttt{ansDepth-cutDepth}})$ from \texttt{edgeList}
\EndIf

\Comment{(v) Generate a formula for each edge, and store in \texttt{forumlaList}}
\State \texttt{formulaList} $\gets [\ ]$
\For{edge $(x_i, x_j)$ in \texttt{edgeList}}
    \State Sample $a, b \in \{-3, -2, -1, 1, 2, 3\}$
    \State Define \texttt{formula} $\gets a \cdot x_i + b \cdot x_j = a \cdot \texttt{varDict}[x_i] + b \cdot \texttt{varDict}[x_j]$
    \State Append \texttt{formula} to \texttt{formulaList} \Comment{So that \texttt{formulaList} has the same order as \texttt{edgeList}}
\EndFor

\Comment{(vi) Adjust the ordering of \texttt{formulaList} according to \texttt{order}}
\If{\texttt{order} $= \text{"backward"}$}
    \State Reverse \texttt{formulaList}
\EndIf
\If{\texttt{order} $= \text{"random"}$}
    \State Random Shuffle \texttt{formulaList}
\EndIf

\Return \texttt{formulaList} \Comment{Formulas serving as conditions of the problem.}
\end{algorithmic}
\end{algorithm}
}

Algorithm \ref{alg:gen_question} generates \texttt{formulaList}, which contains the formulas that will serve as the conditions of the problem.
To translate that into natural language, item names will be sampled according to the \textit{compositeName} option. Then, \texttt{formulaList} can be translated to natural language using pre-defined templates. The question sentence will simply be ``what is the price of \texttt{\{item name of the questioned variable\}}''.

We want to point out that although all the variables are assigned a value in \texttt{varDict}, this is purely for the sake of (i) subsequently generating the random formulas (ii) guaranteeing that all calculable variables will have values between 5 and 15. When \texttt{unanswerable} $=true$, the \textit{cut} will guarantee that the problem is unanswerable.

In the following, we also detail the simple breadth-first traversal algorithm for getting all the edges from the tree, which enables us to control the order of the conditions in the problem.
{\small
\begin{algorithm}[ht!]
\caption{Breadth-First Traversal to Get Edges}
\label{alg:tree}
\begin{algorithmic}[1]
\Require \texttt{root}: the root of a tree

\Comment{Get the list of all edges via a breadth-first traversal}
\State \texttt{edgeList} $\gets [\ ]$, \ \texttt{q} $\gets$ a queue containing \texttt{root}
\While{\texttt{q} is not empty}
    \State \texttt{node} $\gets$ \texttt{q.dequeue()}
    \For{\texttt{child} $\in$ \texttt{node.children}}
        \State Add \texttt{(node, child)} to \texttt{edgeList}
        \State Add \texttt{child} to \texttt{q}
    \EndFor
\EndWhile

\Return \texttt{edgeList}
\end{algorithmic}
\end{algorithm}
}

Given a \texttt{formulaList}, each formula is translated via the following template:
\begin{tcolorbox}[colback=gray!10, colframe=gray!90, title=Formula Translation Template, fontupper=\ttfamily]
Suppose $x_1$ stands for <dish1> at <restaurant1>, and $x_2$ stands for <dish2> at <restaurant2>.
\begin{enumerate}
\item If the formula is $a_1 x_1 + a_2 x_2 = b$ where $a_1,a_2>0$, then the formula will be translated to: ``$a_1$ <dish1> at <restaurant1> and $a_2$ <dish2> at <restaurant2> cost $b$ dollars''.
\item If the formula is $a_1 x_1 - a_2 x_2 = 0$ where $a_1,a_2>0$, then the formula will be translated to: ``The price of $a_1$ <dish1> at <restaurant1> is the same as the price of $a_2$ <dish2> at <restaurant2>''.
\item If the formula is $a_1 x_1 - a_2 x_2 = b$ where $a_1,a_2>0$ and $b>0$, then the formula will be translated to either ``$a_1$ <dish1> at <restaurant1> cost $b$ dollars more than $a_2$ <dish2> at <restaurant2>'' or ``$a_2$ <dish2> at <restaurant2> cost $b$ dollars less than $a_1$ <dish1> at <restaurant1>''.
\item The case of $b<0$ is similar to the case above.
\end{enumerate}
\end{tcolorbox}

\section{Details of Experiments} \label{a:exp}

\subsection{Prompt Template}
Below is the prompt template used for evaluation, which is a 0-shot prompting template with chain-of-thought system message. For reasoning models like o3-mini, we do not use the chain-of-thought system message.
\begin{tcolorbox}[colback=gray!10, colframe=gray!90, title=Zero-shot Evaluation Prompt Template, fontupper=\ttfamily]
// system message (for non-reasoning models) \\
As an expert problem solver, solve step by step the following mathematical questions. \\

// user message \\
Please solve the following math question, and then answer in the form 'Answer: x'. If the known conditions are not sufficient to answer the question, please answer in the form 'Answer: unknown.'. \\

Question: \{\{question\}\} \\

Your solution: 
\end{tcolorbox}

\subsection{Evaluation}
For all the models other than o3-mini, we set \texttt{\{max\_tokens=4000, temperature=0\}}. o3-mini does not allow setting \texttt{temperature}, and we set \texttt{\{max\_completion\_tokens=32000, reasoning\_effort=``high''\}}. GPT-4o and o3-mini are evaluated via calling OpenAI API. The exact version numbers are ``gpt-4o-2024-08-06'' and ``o3-mini-2025-01-31'', respectively.

When evaluating model's answer to an unanswerable problem, we change the response to all lower case, and if ``unknown'' is mentioned after the last occurrence of the trigger word ``answer'', it is counted as a correct answer.

\section{Examples of Hallucinatory Answers}
\subsection{GPT-4o}

\paragraph{Example 1:}

A BLT sandwich at Urban Plate costs 13 dollars. 3 burgers at Taste Good Cuisine cost 6 dollars more than 3 fruit tarts at Taste Good Cuisine. A BLT sandwich at Taste Good Cuisine costs 5 dollars. A fruit tart at Taste Good Cuisine costs 9 dollars less than 3 burgers at Urban Plate. 2 burgers at Taste Good Cuisine cost 8 dollars more than a fruit tart at Urban Plate. Question: how much does a burger at Taste Good Cuisine cost?

\noindent\textbf{Model answer:}

{\small
\begin{verbatim}
To solve this problem, we need to set up equations based on the information given and solve for the unknowns.

Let's define the variables:
- Let \( B_T \) be the cost of a burger at Taste Good Cuisine.
- Let \( F_T \) be the cost of a fruit tart at Taste Good Cuisine.
- Let \( B_U \) be the cost of a burger at Urban Plate.
- Let \( F_U \) be the cost of a fruit tart at Urban Plate.

We are given the following information:

1. A BLT sandwich at Urban Plate costs 13 dollars.
2. 3 burgers at Taste Good Cuisine cost 6 dollars more than 3 fruit tarts at Taste Good Cuisine.
   \[
   3B_T = 3F_T + 6
   \]
3. A BLT sandwich at Taste Good Cuisine costs 5 dollars.
4. A fruit tart at Taste Good Cuisine costs 9 dollars less than 3 burgers at Urban Plate.
   \[
   F_T = 3B_U - 9
   \]
5. 2 burgers at Taste Good Cuisine cost 8 dollars more than a fruit tart at Urban Plate.
   \[
   2B_T = F_U + 8
   \]

We need to find the cost of a burger at Taste Good Cuisine, \( B_T \).

From equation (2), we can express \( F_T \) in terms of \( B_T \):
\[
F_T = B_T - 2
\]

Substitute this expression for \( F_T \) into equation (4):
\[
B_T - 2 = 3B_U - 9
\]
\[
B_T = 3B_U - 7
\]

Now, let's substitute \( F_T = B_T - 2 \) into equation (5):
\[
2B_T = (B_T - 2) + 8
\]
\[
2B_T = B_T + 6
\]
\[
B_T = 6
\]

Thus, the cost of a burger at Taste Good Cuisine is 6 dollars.

Answer: 6
\end{verbatim}
}

\noindent\textbf{Mistake:} When the model did ``\texttt{Now, let's substitute F\_T = B\_T - 2 into equation (5)}'', it actually confused \texttt{F\_T} with \texttt{F\_U} in \texttt{equation (5)}. 

\paragraph{Example 2:}

3 Greek salads at Texas BBQ cost 6 dollars less than 3 pieces of cheese cake at Texas BBQ. 3 burgers at Texas BBQ and 2 pieces of cheese cake at Texas BBQ cost 59 dollars. 3 burgers at Taste Good Cuisine and 2 Greek salads at Taste Good Cuisine cost 42 dollars. 2 burgers at Texas BBQ and a piece of cheese cake at Taste Good Cuisine cost 32 dollars. 3 burgers at Texas BBQ cost 15 dollars more than 3 fruit tarts at Taste Good Cuisine. A burger at Taste Good Cuisine costs 10 dollars. A fruit tart at Texas BBQ costs 2 dollars more than a Greek salad at Texas BBQ. Question: how much does a Greek salad at Texas BBQ cost?

\noindent\textbf{Model answer:}

{\small
\begin{verbatim}
To solve this problem, we need to set up equations based on the information given and solve for the cost of 
a Greek salad at Texas BBQ. Let's define the variables:

- \( G \): Cost of a Greek salad at Texas BBQ
- \( C \): Cost of a piece of cheesecake at Texas BBQ
- \( B \): Cost of a burger at Texas BBQ
- \( F \): Cost of a fruit tart at Taste Good Cuisine
- \( B_T \): Cost of a burger at Taste Good Cuisine (given as $10)
- \( F_T \): Cost of a fruit tart at Texas BBQ

Now, let's translate the information into equations:

1. "3 Greek salads at Texas BBQ cost 6 dollars less than 3 pieces of cheesecake at Texas BBQ."
   \[
   3G = 3C - 6
   \]

2. "3 burgers at Texas BBQ and 2 pieces of cheesecake at Texas BBQ cost 59 dollars."
   \[
   3B + 2C = 59
   \]

3. "3 burgers at Taste Good Cuisine and 2 Greek salads at Taste Good Cuisine cost 42 dollars."
   \[
   3B_T + 2G = 42
   \]
   Since \( B_T = 10 \), substitute:
   \[
   3(10) + 2G = 42 \implies 30 + 2G = 42 \implies 2G = 12 \implies G = 6
   \]

4. "2 burgers at Texas BBQ and a piece of cheesecake at Taste Good Cuisine cost 32 dollars."
   \[
   2B + C_T = 32
   \]
   (Note: We don't have enough information about \( C_T \), the cost of cheesecake at Taste Good Cuisine,
   so we can't use this equation directly.)

5. "3 burgers at Texas BBQ cost 15 dollars more than 3 fruit tarts at Taste Good Cuisine."
   \[
   3B = 3F + 15
   \]

6. "A fruit tart at Texas BBQ costs 2 dollars more than a Greek salad at Texas BBQ."
   \[
   F_T = G + 2
   \]

Now, let's solve the equations:

From equation 3, we found \( G = 6 \).

Let's verify if this value is consistent with other equations:

- From equation 1:
  \[
  3G = 3C - 6 \implies 3(6) = 3C - 6 \implies 18 = 3C - 6 \implies 3C = 24 \implies C = 8
  \]

- From equation 2:
  \[
  3B + 2C = 59 \implies 3B + 2(8) = 59 \implies 3B + 16 = 59 \implies 3B = 43 \implies B = \frac{43}{3}
  \]
  (This value seems unusual, but let's proceed.)

- From equation 5:
  \[
  3B = 3F + 15 \implies \frac{43}{3} = F + 5 \implies F = \frac{43}{3} - 5 = \frac{43}{3} - \frac{15}{3} = \frac{28}{3}
  \]

- From equation 6:
  \[
  F_T = G + 2 \implies F_T = 6 + 2 = 8
  \]

The calculations are consistent with the given conditions. Therefore, the cost of a Greek salad at Texas BBQ is:

Answer: 6
\end{verbatim}
}

\noindent\textbf{Mistake:} At step 3, model calculated the value of ``$G$'', but from the start of the solution, ``$G$'' denotes the ``cost of a Greek salad at Texas BBQ'', not the cost of a Greek salad at Taste Good Cuisine.

\subsection{o3-mini} \label{sec:o3-mini}

\paragraph{Example 1:}

A pizza at Taste Good Cuisine and 3 lasagnas at Taste Good Cuisine cost 48 dollars. A lasagna at Urban Plate costs 6 dollars. 3 lasagnas at Urban Plate and 3 pizzas at Urban Plate cost 63 dollars. Question: how much does a lasagna at Taste Good Cuisine cost?

\noindent\textbf{Model answer:}

{\small
\begin{verbatim}
Let p be the price of a pizza at Taste Good Cuisine and l be the price of a lasagna at Taste Good Cuisine.
We are given:

1. p + 3l = 48.                   (Taste Good Cuisine)
2. A lasagna at Urban Plate costs 6 dollars.
3. 3 lasagnas at Urban Plate and 3 pizzas at Urban Plate cost 63 dollars.

From (3), the total cost of 3 lasagnas at Urban Plate is 3 × 6 = 18 dollars. This means that the 3 pizzas at
Urban Plate cost 63 - 18 = 45 dollars, so one pizza at Urban Plate costs 45/3 = 15 dollars.

A common (though unstated) assumption in these puzzles is that dishes with the same name cost the same across 
restaurants. Thus, we assume that a pizza at Taste Good Cuisine also costs 15 dollars. Substitute p = 15 
into equation (1):

15 + 3l = 48  
3l = 48 - 15 = 33  
l = 33/3 = 11

Answer: 11
\end{verbatim}
}

\noindent\textbf{Mistake:} The model made an assumption: ``\texttt{A common (though unstated) assumption in these puzzles is that dishes with the same name cost the same across 
restaurants.}'' However, this cannot be a mathematically valid assumption because following this assumption, it arrives at the conclusion that the price for lasagna at Urban Plate is not the same as that at Taste Good Cuisine (6 dollars versus 11 dollars).

We randomly selected 20 problems from the $\texttt{ansDepth}=2$ set that triggered hallucinatory answer from o3-mini and manually checked the model answer, all of them shares this same mistake. Below is a similar example:

\paragraph{Example 2:}

A scrambled egg at Bistro Nice and a piece of cheese cake at Mike's Place cost 21 dollars. 2 scrambled eggs at Bistro Nice cost 6 dollars more than 2 pieces of cheese cake at Bistro Nice. A scrambled egg at Mike's Place costs 6 dollars. Question: how much does a scrambled egg at Bistro Nice cost?

\noindent\textbf{Model answer:}

{\small
\begin{verbatim}
Let S_B = price of a scrambled egg at Bistro Nice,
    C_B = price of a piece of cheesecake at Bistro Nice,
    C_M = price of a piece of cheesecake at Mike’s Place.

The problem gives:
1. S_B + C_M = 21 dollars.
2. Two scrambled eggs at Bistro Nice cost 6 dollars more than two cheesecakes at Bistro Nice:
  2S_B = 2C_B + 6  →  S_B = C_B + 3.
3. A scrambled egg at Mike’s Place costs 6 dollars (this is extra information).

Without further assumptions the system involves three unknowns (S_B, C_B, C_M) and only two equations. 
However, it is common in these problems to assume that when the same item is mentioned (here “cheesecake”) 
it has a consistent price from one restaurant to the other. That is, we assume C_M = C_B.

Under that assumption the first equation becomes:
  S_B + C_B = 21.
Then using S_B = C_B + 3 we substitute:
  (C_B + 3) + C_B = 21  →  2C_B + 3 = 21  →  2C_B = 18  →  C_B = 9.
Thus, S_B = 9 + 3 = 12.

Answer: 12
\end{verbatim}
}

\noindent\textbf{Mistake:} Same as above.

\paragraph{Example 3:}

3 BLT sandwiches at Bistro Nice cost 6 dollars more than 3 pieces of cheese cake at Urban Plate. 3 burgers at Urban Plate cost 18 dollars less than 3 pieces of cheese cake at Bistro Nice. 2 BLT sandwiches at Urban Plate and 2 scrambled eggs at Urban Plate cost 34 dollars. A BLT sandwich at Urban Plate costs 8 dollars. 3 pieces of cheese cake at Bistro Nice cost 12 dollars more than 3 burgers at Bistro Nice. 2 burgers at Urban Plate cost 10 dollars less than 2 scrambled eggs at Bistro Nice. 3 pieces of cheese cake at Bistro Nice and a piece of cheese cake at Urban Plate cost 54 dollars. Question: how much does a piece of cheese cake at Urban Plate cost?

\noindent\textbf{Model answer:}

{\small
\begin{verbatim}
Let CC_U be the cost of one piece of cheese cake at Urban Plate.

Step 1. From “3 BLT sandwiches at Bistro Nice cost 6 dollars more than 3 pieces of cheese cake at Urban Plate”:
  3·BLT_B = 3·CC_U + 6  →  BLT_B = CC_U + 2     (1)

Step 2. From “3 burgers at Urban Plate cost 18 dollars less than 3 pieces of cheese cake at Bistro Nice”:
  3·BUR_U = 3·CC_B - 18  →  With BUR_U determined later, we use this after finding BUR_U.

Step 3. “2 BLT sandwiches at Urban Plate and 2 scrambled eggs at Urban Plate cost 34 dollars”:
  2·BLT_U + 2·Eggs_U = 34  and we are given in Step 4 that BLT_U = 8.
  Thus, 2·8 + 2·Eggs_U = 34  →  16 + 2·Eggs_U = 34  →  2·Eggs_U = 18  →  Eggs_U = 9.

Step 4. “A BLT sandwich at Urban Plate costs 8 dollars”:
  BLT_U = 8.

Step 5. “3 pieces of cheese cake at Bistro Nice cost 12 dollars more than 3 burgers at Bistro Nice”:
  3·CC_B = 3·BUR_B + 12  →  CC_B = BUR_B + 4     (2)

Step 6. “2 burgers at Urban Plate cost 10 dollars less than 2 scrambled eggs at Urban Plate”:
  2·BUR_U = 2·Eggs_U - 10  →  2·BUR_U = 18 - 10  →  2·BUR_U = 8  →  BUR_U = 4.

Step 7. Now return to the burgers at Urban Plate in Step 2:
  3·BUR_U = 3·CC_B - 18  With BUR_U = 4,
  3·4 = 3·CC_B - 18  →  12 = 3·CC_B - 18  →  3·CC_B = 30  →  CC_B = 10.

Step 8. “3 pieces of cheese cake at Bistro Nice and a piece of cheese cake at Urban Plate cost 54 dollars”:
  3·CC_B + CC_U = 54  →  3·10 + CC_U = 54  →  30 + CC_U = 54  →  CC_U = 24.

The question asks: “how much does a piece of cheese cake at Urban Plate cost?”

Answer: 24
\end{verbatim}
}

\noindent\textbf{Mistake:} \texttt{Step 6. “2 burgers at Urban Plate cost 10 dollars less than 2 scrambled eggs at Urban Plate”.} That is a hallucinatory condition. The actual condition is "2 burgers at Urban Plate cost 10 dollars less than 2 scrambled eggs at Bistro Nice".

\end{document}